\pdfoutput=1
\documentclass[11pt]{article}
\usepackage{emnlp2021}
\usepackage{latexsym}
\usepackage{newtxtext}
\usepackage{newtxmath}
\usepackage{framed}

\usepackage{soul}

\usepackage{microtype}

\usepackage{amsmath,amsfonts,bm}

\def\figref#1{figure~\ref{#1}}

\def\secref#1{section~\ref{#1}}

\def\eqref#1{equation~\ref{#1}}

\def\1{\bm{1}}

\DeclareMathAlphabet{\mathsfit}{\encodingdefault}{\sfdefault}{m}{sl}
\SetMathAlphabet{\mathsfit}{bold}{\encodingdefault}{\sfdefault}{bx}{n}

\usepackage{hyperref}
\usepackage{url}
\usepackage{wrapfig}

\usepackage{latexsym}
\usepackage[ruled,vlined]{algorithm2e}
\usepackage{pgfplots}
\usepackage{verbatim}
\usepackage{mathtools}

\usetikzlibrary{patterns}

\usepackage{microtype}

\usepackage{xcolor}

\usepackage{booktabs}

\usepackage{multirow}
\usepackage{subfigure}

\title{Answering Open-Domain Questions of Varying Reasoning Steps from Text}

\author{Peng Qi*$^{\spadesuit\heartsuit}$\qquad Haejun Lee*$^{\clubsuit}$ \qquad Oghenetegiri ``TG'' Sido*$^{\spadesuit}$\qquad Christopher D. Manning$^{\spadesuit}$ \\
 $\spadesuit$ Computer Science Department, Stanford University \\
 $\heartsuit$ JD AI Research \\
 $\clubsuit$ Samsung Research \\
  \texttt{\{pengqi, osido, manning\}@cs.stanford.edu, haejun82.lee@samsung.com} }

\date{}

\definecolor{red}{HTML}{E31A1C}
\definecolor{blue}{HTML}{1F78B4}
\definecolor{green}{HTML}{33A02C}
\definecolor{orange}{HTML}{FF7F00}
\definecolor{purple}{HTML}{6A3D9A}

\newcommand{\ie}{\textit{i.e.}}
\newcommand{\eg}{\textit{e.g.}}

\newcommand{\deleted}[1]{\textcolor{red}{ \ul{#1}}}
\newcommand{\modified}[1]{\textcolor{orange}{ \textit{#1}}}
\newcommand{\added}[1]{\textcolor{green}{ \textbf{#1}}}

\newcommand{\fone}{F\textsubscript{1}}

\newcommand{\squad}{SQuAD}
\newcommand{\squadopen}{\squad{} Open}
\newcommand{\hotpotqa}{HotpotQA}
\newcommand{\triviaqa}{TriviaQA}

\newcommand{\beerqa}{\textsc{BeerQA}}

\newcommand{\golden}{\textsc{GoldEn}}
\newcommand{\irrr}{IRRR}

\newcommand{\clstoken}{\texttt{[CLS]}}
\newcommand{\septoken}{\texttt{[SEP]}}
\newcommand{\conttoken}{\texttt{[CONT]}}

\newif\ifmarginalnotes
\marginalnotestrue
\definecolor{CMpurple}{rgb}{0.6,0.18,0.64}

\renewcommand{\figref}[1]{Figure~\ref{#1}}

\pgfplotsset{width=7.5cm, height=5.5cm,compat=1.16,every axis plot/.append style={
line width=.75pt,}}

\begin{document}
\maketitle

\newif\ifaclfinal
\aclfinaltrue
\ifaclfinal
\renewcommand*{\thefootnote}{\fnsymbol{footnote}}
\setcounter{footnote}{1}
\footnotetext{These authors contributed equally.}
\renewcommand*{\thefootnote}{\arabic{footnote}}
\setcounter{footnote}{0}
\fi

\begin{abstract}

    We develop a unified system to answer directly from text open-domain questions that may require a varying number of retrieval steps.
    We employ a single multi-task transformer model to perform all the necessary subtasks---retrieving supporting facts, reranking them, and predicting the answer from all retrieved documents---in an iterative fashion.
    We avoid crucial assumptions of previous work that do not transfer well to real-world settings, including exploiting knowledge of the fixed number of retrieval steps required to answer each question or using structured metadata like knowledge bases or web links that have limited availability.
    Instead, we design a system that can answer open-domain questions  on any text collection without prior knowledge of reasoning complexity.
    To emulate this setting, we construct a new benchmark, called \beerqa{}, by combining existing one- and two-step datasets with a new collection of 530 questions that require three Wikipedia pages to answer, unifying Wikipedia corpora versions in the process.
    We show that our model demonstrates competitive performance on both existing benchmarks and this new benchmark.
    We make the new benchmark available at \url{https://beerqa.github.io/}.

\end{abstract}

\section{Introduction}

Using knowledge to solve problems is a hallmark of intelligence.
Since human knowledge is often containned in large text collections, open-domain question answering (QA) is an important means for intelligent systems to make use of the knowledge in large text collections.
With the help of large-scale datasets based on Wikipedia \citep{rajpurkar2016squad, rajpurkar2018know} and other large corpora \citep{trischler2016newsqa, dunn2017searchqa, talmor2018web}, the research community has made substantial progress on tackling this problem in recent years, including in the direction of complex reasoning over multiple pieces of evidence, or \emph{multi-hop} reasoning \citep{yang2018hotpotqa, welbl2018constructing, chen-etal-2020-hybridqa}. 

Despite this success, most previous systems are developed with, and evaluated on, datasets that contain exclusively \emph{single-hop} questions (ones that require a single document or paragraph to answer) or \emph{two-hop} ones.
As a result, their design is often tailored exclusively to single-hop \citep[\eg,][]{chen2017reading, wang2018evidence} or multi-hop questions \citep[\eg,][]{nie2019revealing, min-etal-2019-multi, feldman2019multi, zhao2020complex, xiong2021answering}. 
Even when the model is designed to work with both, it is often trained and evaluated on exclusively single-hop or multi-hop settings \citep[\eg,][]{asai2020learning}.
In practice, not only can we not expect open-domain QA systems to receive exclusively single- or multi-hop questions from users, but it is also non-trivial to judge reliably whether a question requires one or multiple pieces of evidence to answer \emph{a priori}.
For instance, \emph{``In which U.S. state was Facebook founded?''} appears to be single-hop, but its answer cannot be found in the main text of a single English Wikipedia page.

Besides the impractical assumption about reasoning hops, previous work often also assumes access to non-textual metadata such as knowledge bases, entity linking, and Wikipedia hyperlinks when retrieving supporting facts, especially in answering complex questions \citep{nie2019revealing,
feldman2019multi, zhao2019transformer, asai2020learning, dhingra2020differentiable, zhao2020complex}.
While this information is helpful, it is not always available in text collections we might be interested in getting answers from, such as news or academic research articles, besides being labor-intensive and time-consuming to collect and maintain.
It is therefore desirable to design a system that is capable of extracting knowledge from text without using such metadata, to maximally emphasize using knowledge available to us in the form of text.

\begin{figure*}
\centering
\includegraphics[width=0.98\textwidth]{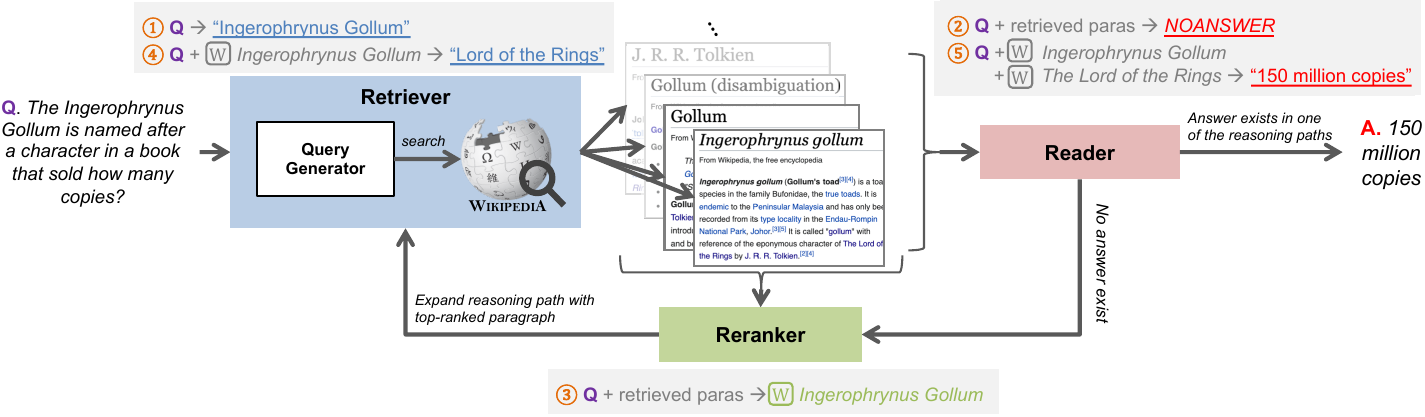} 
\caption{The \irrr{} question answering pipeline answers a complex question in the \hotpotqa{} dataset by iteratively retrieving, reading, and reranking paragraphs from Wikipedia.
In this example, the question is answered in five steps: 1. the retriever model selects the words ``Ingerophrynus gollum'' from the question as an initial search query; 2. the question answering model attempts to answer the question by combining the question with each of the retrieved paragraphs and fails to find an answer; 3. the reranker picks the paragraph about the \emph{Ingerophrynus gollum} toad to extend the reasoning path; 4. the retriever generates an updated query ``Lord of the Rings'' to retrieve new paragraphs; 5. the reader correctly predicts the answer ``150 million copies'' by combining the reasoning path (question + ``Ingerophrynus gollum'') with the newly retrieved paragraph about ``The Lord of the Rings''.
}\label{fig:pipeline}
\end{figure*}

To address these limitations, we propose Iterative Retriever, Reader, and Reranker (\irrr{}), which features a single neural network model that performs all of the subtasks required to answer questions from a large collection of text (see Figure \ref{fig:pipeline}).
\irrr{} is designed to leverage off-the-shelf information retrieval systems by generating natural language search queries, which allows it to easily adapt to arbitrary collections of text without requiring well-tuned neural retrieval systems or extra metadata.
This further allows users to understand and control \irrr{}, if necessary, to facilitate trust.
Moreover, \irrr{} iteratively retrieves more context to answer the question, which allows it to easily accommodate questions of different number of reasoning steps.

To evaluate the performance of open-domain QA systems in a more realistic setting, we construct a new benchmark called \beerqa{}\footnote{\url{https://beerqa.github.io/}} by combining the questions from the single-hop \squadopen{} \citep{rajpurkar2016squad, chen2017reading} and the two-hop \hotpotqa{} \citep{yang2018hotpotqa} with a new collection of 530 human-annotated questions that require information from at least three Wikipedia pages to answer.
We map all questions to a unified version of the English Wikipedia to reduce stylistic differences that might provide statistical shortcuts to models.
As a result, \beerqa{} provides a more realistic evaluation of open-ended question answering systems in their ability to answer questions without knowledge of the number of reasoning steps required ahead of time.
We show that \irrr{} not only achieves competitive performance with state-of-the-art models on the original \squadopen{} and \hotpotqa{} datasets, but also establishes a strong baseline for this new dataset.

To recap, our contributions in this paper are: (1)
a new open-domain QA benchmark, \beerqa, that features questions requiring variable steps of reasoning to answer on a unified Wikipedia corpus.
(2) A single unified neural network model that performs all essential subtasks in open-domain QA purely from text (retrieval, reranking, and reading comprehension), which not only achieves strong results on \squad{} and \hotpotqa{}, but also establishes a strong baseline on this new benchmark.%
\footnote{Our code for the model can be found at: \url{https://github.com/beerqa/IRRR}.}

\section{Open-Domain Question Answering}

The task of open-domain question answering is concerned with finding the answer $a$ to a question $q$ from a large text collection $\mathcal D$.
Successful solutions to this task usually involve two crucial components: an \emph{information retrieval} system that finds a small set of relevant documents $\mathcal D_r$ from $\mathcal D$, and a \emph{reading comprehension} system that extracts the answer from it.%
\footnote{Some recent work breaks away from this mold, and use large pretrained language models \citep[\eg, T5; ][]{raffel2020t5} to directly generate answers from knowledge stored in model parameters.}
\citet{chen2017reading} presented the first neural-network-based approach to this problem, which was later extended by \citet{wang2018reinforced} with a \emph{reranking} system to further reduce the amount of context the reading comprehension component has to consider to improve answer accuracy.

More recently, \citet{yang2018hotpotqa} showed that this single-step retrieve-and-read approach to open-domain question answering is inadequate for more complex questions that require multiple pieces of evidence to answer (\eg, \emph{``What is the population of Mark Twain's hometown?''}).
While later work approaches these by extending supporting fact retrieval beyond one step, most assumes that all questions are either exclusively single-hop or multi-hop during training and evaluation.
We propose \irrr{}, a system that performs variable-hop retrieval for open-domain QA to address these issues, and present a new benchmark, \beerqa{}, to evaluate systems in a more realistic setting.

\section{\irrr{}: Iterative Retriever, Reader, and Reranker}

In this section, we present a unified model to perform all of the subtasks necessary for open-domain question answering---Iterative Retriever, Reader, and Reranker (\irrr{}), which performs the subtasks involved in an iterative manner to accommodate questions with a varying number of steps.
\irrr{} aims at building a reasoning path $p$ from the question $q$, through all the necessary supporting documents or paragraphs $d\in \mathcal{D}_{\text{gold}}$ to the answer $a$ (where $\mathcal{D}_{\text{gold}}$ is the set of gold supporting facts).%
\footnote{For simplicity, we assume that there is a single set of relevant supporting facts that helps answer each question.}
As shown in \figref{fig:pipeline}, \irrr{} operates in a loop of retrieval, reading, and reranking to expand the reasoning path $p$ with new documents from $d\in \mathcal{D}$.

Specifically, given a question $q$, we initialize the reasoning path with the question itself, \ie, $p_0 = [q]$, and generate from it a search query with \irrr{}'s retriever.
Once a set of relevant documents $\mathcal{D}_1 \subset \mathcal{D}$ is retrieved, they might either help answer the question, or reveal clues about the next piece of evidence we need to answer $q$.
The reader model then attempts to read each of the documents in $\mathcal{D}_1$ to answer the question combined with the current reasoning path $p$.
If more than one answer can be found from these candidate reasoning paths, we predict the answer with the highest \emph{answerability} score, which we will detail in \secref{sec:reader}.
If no answer can be found, then \irrr{}'s reranker scores each retrieved paragraph against the current reasoning path, and appends the top-ranked paragraph to the current reasoning path, \ie, $p_{i+1} = p_i + [\arg\max_{d \in D_1} \mathrm{reranker}(p_i, d)]$, before the updated reasoning path is presented to the retriever to generate new search queries.
This iterative process is repeated until an answer is predicted from one of the reasoning paths, or until the reasoning path has reached a cap of $K$ documents.%

\begin{figure}
\centering
\includegraphics[width=.48\textwidth]{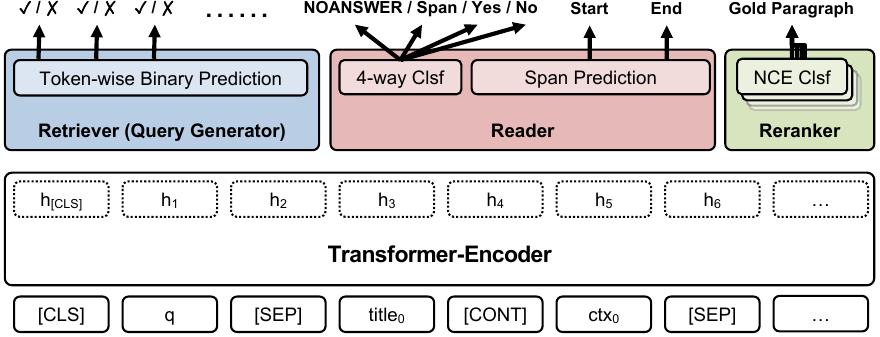} 
\caption{The overall architecture of our IRRR model, which uses a shared Transformer encoder to perform all subtasks of open-domain question answering. }\label{fig:irrr_arch}
\end{figure}

To reduce computational cost and improve model representations of reasoning paths from shared statistical learning, %
\irrr{} is implemented as a multi-task model built on a pretrained Transformer model that performs all three subtasks.
At a high level, it consists of a Transformer encoder \citep{vaswani2017attention} which takes the reasoning path $p$ (the question and all retrieved paragraphs so far) as input, and one set of task-specific parameters for each task of retrieval, reranking, and reading comprehension (see \figref{fig:irrr_arch}).
The retriever generates natural language search queries by selecting words from the reasoning path, the reader extracts answers from the reasoning path and abstains if its confidence is not high enough, and the reranker assigns a scalar score for each retrieved paragraph as a potential continuation of the current reasoning path.

The input to our Transformer encoder is formatted similarly to that of the BERT model \citep{devlin2019bert}. 
For a reasoning path $p$ that consists of the question and $t$ retrieved paragraphs, the input is formatted as ``\clstoken{} question \septoken{} title\textsubscript{1} \conttoken{} para\textsubscript{1}%
\septoken{} \ldots\ title\textsubscript{$t$} \conttoken{} para\textsubscript{$t$} \septoken{}'', where \clstoken{}, \septoken{}, and \conttoken{} are special tokens to separate different components of the input.
The \conttoken{} embedding is randomly initialized with a truncated normal distribution with a standard deviation of 0.02, and finetuned with other model parameters during training.

We will detail each of the task-specific components in the following subsections.

\subsection{Retriever}

The goal of the retriever is to generate natural language queries to retrieve relevant documents from an off-the-shelf text-based retrieval engine.%
\footnote{We employ Elasticsearch \citep{gormley2015elasticsearch} as our text-based search engine, and follow previous work to process Wikipedia and search results, which we detail in Appendix \ref{sec:search_engine}.}
This allows \irrr{} to perform open-domain QA in an explainable and controllable manner, where a user can easily understand the model's behavior and intervene if necessary.
We extract search queries from the current reasoning path, \ie, the original question and all of the paragraphs that we have already retrieved, similar to \golden{} Retriever's approach \citep{qi2019answering}.
This is based on the observation that there is usually a strong semantic overlap between the reasoning path and the next paragraph to retrieve, which helps reduce the search space of potential queries.
We note, though, that \irrr{} differs from \golden{} Retriever in two important ways:
(1) we allow search queries to be any subsequence of the reasoning path instead of limiting it to substrings to allow for more flexible combinations of search phrases; (2) more importantly, we employ the same retriever model across reasoning steps to generate queries instead of training separate ones for each reasoning step, which is crucial for \irrr{} to generalize to arbitrary reasoning steps.

To predict these search queries from the reasoning path, we apply a token-wise binary classifier on top of the shared Transformer encoder model, to decide whether each token is included in the final query.
At training time, we derive supervision signal to train these classifiers with a binary cross entropy loss (which we detail in Section \ref{sec:query_oracle}); at test time, we select a cutoff threshold for query words to be included from the reasoning path.
In practice, we find that boosting the model to predict more query terms is beneficial to increase the recall of the target paragraphs in retrieval.

\subsection{Reader} \label{sec:reader}

The reader model attempts to find the answer given a reasoning path comprised of the question and retrieved paragraphs. 
To support unanswerable questions and the special non-extractive answers  \emph{yes} and \emph{no} from \hotpotqa{}, we train a classifier conditioned on the Transformer encoder representation of the \clstoken{} token to predict one of the 4 classes \texttt{SPAN}\slash \texttt{YES}\slash\texttt{NO}\slash\texttt{NOANSWER}. The classifier thus simultaneously assigns an \emph{answerability} score to this reasoning path to assess the likelihood of the document having the answer to the original question on this reasoning path.
Span answers are predicted from the context using a span start classifier and a span end classifier, following \citet{devlin2019bert}.

We define answerability as the log likelihood ratio between the most likely positive answer and the \texttt{NOANSWER} prediction, and use it to pick the best answer from all the candidate reasoning paths to stop \irrr{}'s iterative process, if found. 
We find that this likelihood ratio formulation is less affected by sequence length compared to prediction probability, thus making it easier to assign a global threshold across reasoning paths of different lengths to stop further retrieval.
We include further details about answerability calculation in Appendix \ref{sec:further_training_details}.

\subsection{Reranker}

When the reader fails to find an answer from the reasoning path, the reranker selects one of the retrieved paragraphs to expand it, so that the retriever can generate new search queries to retrieve new context to answer the question.
To achieve this, we assign each potential extended reasoning path a score by linearly transforming the hidden representation of the \clstoken{} token, and picking the extension that has the highest score.
At training time, we normalize the reranker scores across top retrieved paragraphs with softmax, and maximize the log likelihood of selecting gold supporting paragraphs from retrieved ones, which is a noise contrastive estimation \citep[NCE;][]{mnih2013learning, jean2015using} of the reranker likelihood over all retrieved paragraphs.

\subsection{Training IRRR}

\subsubsection{Dynamic Oracle for Query Generation} \label{sec:query_oracle}

Since existing open-domain QA datasets do not include human-annotated search queries, we need to derive supervision signal to train the retriever with a dynamic oracle.
Similar to \golden{} Retriever, %
we derive search queries from overlapping terms between the reasoning path and the target paragraph with the goal of maximizing retrieval performance.

To reduce computational cost, we limit our attention to overlapping spans of text between the reasoning path and the target document when generating oracle queries.
For instance, when ``David'' is part of the overlapping span ``David Dunn'', the entire span is either included or excluded from the oracle query to reduce the search space.
Once $N$ overlapping spans are found, we approximate the importance of each with the following ``importance'' metric to avoid enumerating all $2^N$ combinations to generate the oracle query
\begin{align}
    \mathrm{Imp}(s_i) &= \mathrm{Rank}(t, \{s_j\}_{j=1, j\ne i}^N) - \mathrm{Rank}(t, \{s_i\}), \nonumber
\end{align}
where $s_j$ are overlapping spans, and $\mathrm{Rank}(t, S)$ is the rank of target document $t$ in the search result when spans $S$ are used as search queries (the smaller, the closer $t$ is to the top).
Intuitively, the second term captures the importance of the search term when used alone, and the first captures its importance when combined with all other overlapping spans, which helps us capture query terms that are only effective when combined.
After estimating importance of each overlapping span, we determine the final oracle query by first sorting all spans by descending importance, then including each in the final oracle query until the search rank of $t$ stops improving. 
The resulting time complexity for generating these oracle queries is thus $O(N)$, \ie, linear in the number of overlapping spans between the reasoning path and the target paragraph.

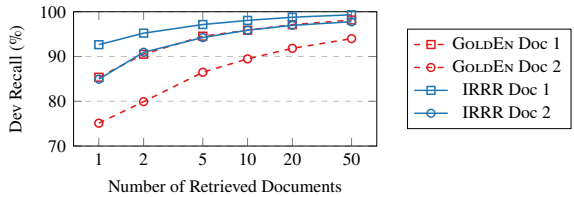
\begin{figure}
    \centering
    \resizebox{\linewidth}{!}{
    \pgfplotstableread[row sep=\\,col sep=&]{
	k & recall \\
	1 & 85.36 \\
	2 & 90.52 \\
	5 & 94.53 \\
	10 & 95.87 \\
	20 & 97.12 \\
	50 & 98.15 \\
}\goldenpone

\pgfplotstableread[row sep=\\,col sep=&]{
	k & recall \\
	1 & 75.11 \\
	2 & 79.91 \\
	5 & 86.47 \\
	10 & 89.47 \\
	20 & 91.82 \\
	50 & 93.99 \\
}\goldenptwo

\pgfplotstableread[row sep=\\,col sep=&]{
	k & recall \\
	1 & 92.64 \\
	2 & 95.21 \\
	5 & 97.14 \\
	10 & 98.06 \\
	20 & 98.77 \\
	50 & 99.32 \\
}\irrrpone

\pgfplotstableread[row sep=\\,col sep=&]{
	k & recall \\
	1 & 84.94 \\
	2 & 90.94 \\
	5 & 94.23 \\
	10 & 95.87 \\
	20 & 96.96 \\
	50 & 97.76 \\
}\irrrptwo

\begin{tikzpicture}[every plot/.append style={thick},font=\footnotesize]
\begin{axis}[
xlabel={Number of Retrieved Documents},
xtick={1, 2, 5, 10, 20, 50},
xticklabels={1,2,5,10,20,50},
xmode=log,
ylabel={Dev Recall (\%)},
width=7cm,
height=4cm,
ymin=70,
ymax=100,
legend style={at={(1.1,0.5)}, anchor=west},
legend columns=1,
ymajorgrids=true,
grid style=dashed,
]
\addplot[dashed, color=red, mark=square, mark options={solid}] plot table[x=k, y=recall]{\goldenpone};
\addplot[dashed, color=red, mark=o, mark options={solid}] plot table[x=k,y=recall]{\goldenptwo};
\addplot[color=blue, mark=square, mark options={solid, fill=blue}] plot table[x=k, y=recall]{\irrrpone};
\addplot[color=blue, mark=o, mark options={solid, fill=blue}] plot table[x=k,y=recall]{\irrrptwo};
\legend{\golden{} Doc 1, \golden{} Doc 2, \irrr{} Doc 1, \irrr{} Doc 2}

\end{axis}
\end{tikzpicture}
    }
    \caption{Recall of the two gold supporting documents by the oracle queries of \golden{} Retriever and \irrr{} on the \hotpotqa{} dataset, where each question corresponds to two supporting documents.}
    \label{fig:oracle_recall}
\end{figure}

Figure \ref{fig:oracle_recall} shows that the added flexibility of non-span queries in \irrr{} significantly improves retrieval performance compared to that of \golden{} Retriever, which is only able to extract contiguous spans from the reasoning path as queries.

\subsubsection{Reducing Exposure Bias with Data Augmentation}\label{sec:data_aug}

With the dynamic oracle, we are able to generate target queries to train the retriever model, retrieve documents to train the reranker model, and expand reasoning paths in the training set by always choosing a gold paragraph, following \citet{qi2019answering}.
However, this might prevent the model from generalizing to cases where model behavior deviates from the oracle.
To address this, we augment the training data by occasionally selecting non-gold paragraphs to expand reasoning paths, and use the dynamic oracle to generate queries for the model to ``recover'' from these synthesized retrieval mistakes.
We found that this data augmentation significantly improves the performance of \irrr{} in preliminary experiments, and thus report main results with augmented training data.

\section{Experiments}

\paragraph{Standard Benchmarks.} We test \irrr{} on two standard benchmarks, \squadopen{} and \hotpotqa{}.
\squadopen{} \cite{chen2017reading} designates the development set of the original \squad{} dataset as its test set, which features more than 10,000 questions, each based on a single paragraph in a Wikipedia article.
For this dataset, we follow previous work and use the 2016 English Wikipedia as the corpus for evaluation.
Since the authors did not present a standard development set, we further split part of the training set to construct a development set roughly as large as the test set.
\hotpotqa{} \citep{yang2018hotpotqa} features more than 100,000 questions that require the introductory paragraphs of two Wikipedia articles to answer, and we focus on its open-domain ``fullwiki'' setting in this work.
For \hotpotqa{}, we use the introductory paragraphs provided by the authors for training and evaluation, which is based on a 2017 Wikipedia dump.

\begin{figure}
    \centering
    \resizebox{\linewidth}{!}{
    \begin{tabular}{p{28em}}
    \toprule
    \textbf{Question:} How many counties are on the island that is home to the fictional setting of the novel in which Daisy Buchanan is a supporting character?\\
    \midrule
    \textbf{Wikipedia Page 1: \textit{Daisy Buchanan}}\\
    Daisy Fay Buchanan is a fictional character in F. Scott Fitzgerald's magnum opus ``The Great Gatsby'' (1925)...\\
    \midrule
    \textbf{Wikipedia Page 2: \textit{The Great Gatsby}}\\
    The Great Gatsby is a 1925 novel ... that follows a cast of characters living in the fictional town of West Egg on prosperous Long Island ...\\
    \midrule
    \textbf{Wikipedia Page 3: \textit{Long Island}}\\
    The Long Island ... comprises four counties in the U.S. state of New York: Kings and Queens ... to the west; and Nassau and Suffolk to the east...\\
    \midrule
    \textbf{Answer:} four\\ 
    \bottomrule
    \end{tabular}
    }
    \caption{An example of the newly collected challenge questions. This particular question requires three pieces of evidence to answer.
    }
    \label{fig:three_hop_example}
\end{figure}

\paragraph{New Benchmark.} To evaluate the performance of \irrr{} as well as future QA systems in a more realistic open-domain setting without a pre-specified number of reasoning steps for each question, we further combine \squadopen{} and \hotpotqa{} with 530 newly collected challenge questions (see Figure \ref{fig:three_hop_example} for an example, and Appendix \ref{sec:three_hop} for more details) to construct a new benchmark.
Note that naively combining the datasets by merging the questions and the underlying corpora is problematic, as the corpora not only feature repeated and sometimes contradicting information, but also make them available in two distinct forms (full Wikipedia pages in one and just the introductory paragraphs in the other).
This could result in models taking corpus style as a shortcut to determine question complexity, or even result in plausible false answers due to corpus inconsistency.

To construct a high-quality unified benchmark, we begin by mapping the paragraphs each question is based on to a more recent version of Wikipedia.\footnote{In this work, we used the English Wikipedia dump from August 1st, 2020.}
We discarded examples where the Wikipedia pages have either been removed or significantly edited such that the answer can no longer be found from paragraphs that are similar enough to the original contexts the questions are based on.\footnote{We refer the reader to Appendix \ref{sec:data_processing} for further details about these Wikipedia corpora and how we process and map between them.}
As a result, we filtered out 22,328 examples from \squad{} Open, and 18,649 examples from \hotpotqa{}'s fullwiki setting.%
\ifaclfinal\else\footnote{We obtained access to the \hotpotqa{} fullwiki test set through personal communications with the authors.}\fi~ We add newly annotated challenge questions to the test set of the new benchmark, which require at least three steps of reasoning to answer.
This allows us to test the generalization capabilities of QA models to this unseen scenario.
The statistics of the final dataset, which we name \beerqa{}, can be found in Table \ref{tab:new_benchmark_stats}.
For all benchmark datasets, we report standard answer exact match (EM) and unigram \fone{} metrics.

\begin{table}
    \centering
    \resizebox{\linewidth}{!}{
    \begin{tabular}{lcccc}
        \toprule
         & \squad{} Open & \hotpotqa{} & 3+ Hop
         & Total  \\
        \midrule
        Train & 59,285 & 74,758 & \phantom{00}0 & 134,043 \\
        Dev & \phantom{0}8,132 & \phantom{0}5,989 & \phantom{00}0 & \phantom{0}14,121\\
        Test & \phantom{0}8,424 & \phantom{0}5,978 & 530 & \phantom{0}14,932 \\
        \midrule
        Total & 75,841 & 86,725 & 530 & 163,096\\
        \bottomrule
    \end{tabular}
    }
    \caption{Counts of QA examples in the new unified benchmark, \beerqa.} 
    \label{tab:new_benchmark_stats}
\end{table}

\paragraph{Training details.} We use ELECTRA\textsubscript{LARGE} \citep{clark2020electra} as the pre-trained initialization for our Transformer encoder. 
We train the model on a combined dataset of \squad{} Open and \hotpotqa{} questions where we optimize the joint loss of the retriever, reader, and reranker components simultaneously in an multi-task learning fashion.
Training data for the retriever and reranker components is derived from the dynamic oracle on the training set of these datasets, where reasoning paths are expanded with oracle queries and by picking the gold paragraphs as they are retrieved for the reader component.
We augment the training data with the technique in Section \ref{sec:data_aug} and expand reasoning paths up to 3 reasoning steps on \hotpotqa{} and 2 on \squadopen{}, and find that this results in a more robust model.
After an initial model is finetuned on this expanded training set, we apply our iterative training technique to further reduce exposure bias of the model by generating more data with the trained model and the dynamic oracle.

\section{Results}

In this section, we present the performance of \irrr{} when evaluated against previous systems on standard benchmarks, and demonstrate its efficacy on our new, unified benchmark, especially with the help of iterative training.

\subsection{Performance on Standard Benchmarks}

We first compare \irrr{} against previous systems on \squad{} Open and the fullwiki setting of \hotpotqa{}. %
On each dataset, we compare the performance of \irrr{} against best previously published systems, as well as unpublished ones on public leaderboards.
For a fairer comparison to previous work, we make use of their respective Wikipedia corpora, and limit the retriever to retrieve 150 paragraphs of text from Wikipedia at each step of reasoning.
We also compare \irrr{} against the Graph Recurrent Retriever \citep[GRR;][]{asai2020learning} on our newly collected 3-hop question challenge test set, using the author's released code and models trained on \hotpotqa{}.
In these experiments, we report \irrr{} performance both from training on the dataset it is evaluated on, and from combining the training data we derived from both \squadopen{} and \hotpotqa{}.

\begin{table}
\small
\centering
    \begin{tabular}{lcc}
        \toprule
        \multirow{2}{5em}{System} & \multicolumn{2}{c}{\squad{} Open} \\
        \cline{2-3}
        & EM & \fone{}\\ 
         \midrule
         DrQA \citep{chen2017reading} & 27.1 & ---\\
         DensePR \citep{karpukhin2020dense}  & 38.1 & --- \\
         BERTserini \citep{yang2019end} & 38.6 & 46.1 \\
         MUPPET \citep{feldman2019multi} & 39.3 & 46.2 \\
         RE\textsuperscript{3} \citep{hu2019retrieve} & 41.9 & 50.2 \\
         Knowledge-aided \citep{zhou2020knowledge} & 43.6 & 53.4 \\
         Multi-passage BERT \citep{wang2019multi}& 53.0 & 60.9 \\
         GRR \citep{asai2020learning}& 56.5 & 63.8\ \\
         FiD \citep{izacard2020leveraging}& 56.7 & --- \\
         SPARTA \citep{zhao2020sparta} & 59.3 & 66.5\\
         \midrule
         \irrr{} (\squad) & 56.8 & 63.2 \\
         \irrr{} (\squad+\hotpotqa) & \textbf{61.8} & \textbf{68.9} \\
         \bottomrule
    \end{tabular}
\caption{End-to-end question answering performance on \squad{} Open, evaluated on the same set of documents as \citet{chen2017reading}. }
\label{tab:benchmark_results_squad}
\end{table}

\begin{table}[t]
\centering
\resizebox{.48\textwidth}{!}{%
    \begin{tabular}{lccp{0em}cc}
        \toprule
        \multirow{2}{5em}{System} & \multicolumn{2}{c}{\hotpotqa{}} && \multicolumn{2}{c}{3+ hop} \\
       
        \cline{2-3} \cline{5-6}
        & EM & \fone{} & & EM & \fone{}\\ 
         \midrule
         GRR  \citep{asai2020learning} & 60.0 & 73.0 && 27.2$^{\dagger}$ & 31.9$^{\dagger}$ \\
         Step-by-step$^\otimes$ & 63.0 & 75.4 && --- & ---\\
         DDRQA \citep{zhang2020ddrqa} & 62.3 & 75.3 && --- & --- \\
         MDR \citep{xiong2021answering} & 62.3 & 75.3&& --- & ---\\
         EBS-SH $^\otimes$ & 65.5 & 78.6 && --- & ---\\
         TPRR $^\otimes$ & 67.0 & 79.5 && --- & --- \\
         HopRetriever \citep{li2020hopretriever} & \textbf{67.1} & \textbf{79.9} && --- & --- \\
         \midrule
         \irrr{} (\hotpotqa) & 65.2 & 78.0 && 29.2\phantom{$^{\dagger}$} & 34.2\phantom{$^{\dagger}$} \\
         \irrr{} (\squad{} + \hotpotqa) & 65.7 & 78.2 && \textbf{32.5}\phantom{$^{\dagger}$} & \textbf{36.7}\phantom{$^{\dagger}$} \\
         \bottomrule
    \end{tabular}
}

\caption{End-to-end question answering performance on \hotpotqa{} and the new 3+ hop challenge questions, evaluated on the official \hotpotqa{} Wikipedia paragraphs. $\otimes$ denotes anonymous/preprint unavailable at the time of writing of this paper. $\dagger$ indicates results we obtained using the publicly available code and pretrained models.}
\vspace{-.5em}
\label{tab:benchmark_results_hotpot_3hop}
\end{table}

As can be seen in Tables \ref{tab:benchmark_results_squad} and \ref{tab:benchmark_results_hotpot_3hop}, \irrr{} achieves competitive performance with previous work, and further outperforms previously published work on \squadopen{} by a large margin when trained on combined data.
It also outperforms systems that were submitted after \irrr{} was initially submitted to the \hotpotqa{} leaderboard.
On the 3+ hop challenge set, we similarly notice a large performance margin between \irrr{} and GRR, although neither is trained with questions requiring three or more hops, demonstrating that \irrr{} generalizes well to questions that require more retrieval steps than the ones seen during training.
We note that the systems that outperform \irrr{} on these datasets typically make use of trainable neural retrieval components, which \irrr{} can potentially benefit from adopting as well.
Specifically, SPARTA \citep{zhao2020sparta} introduces a neural sparse retrieval system that potentially works well with \irrr{}'s oracle query generation procedure to further improve retrieval performance, thanks to its use of natural language queries.
HopRetriever \citep{li2020hopretriever} introduces a novel representation of documents for retrieval that is particularly suitable for discovering documents connected by the same entity to answer multi-hop questions, which \irrr{} could benefit from as well.
We leave exploration of these directions to future work.

\begin{figure}[t]

\centering
\footnotesize
\resizebox{.46\textwidth}{!}{
\hskip-1em
\begin{tikzpicture}
\begin{axis}[
    ybar,
    height=3.5cm,
    bar width=7pt,
    title={\squadopen{}},
    xlabel={Retrieval Steps/Question},
    ylabel={Percentage},
    xtick={1,2,3,4,5},
    xmin=.5,xmax=5.5,
    legend pos=north east,
    ymajorgrids=true,
    grid style=dashed,
]

\addplot+[color=blue,fill=blue,postaction={pattern=crosshatch dots,pattern color=blue!40!white}] coordinates {(1,70.338)(2,2.53)(3,12.899)(4,8.837)(5,5.3938)};

\addplot+[color=green,fill=green,postaction={pattern=horizontal lines,pattern color=green!40!white}] coordinates {(1,73.544)(2,2.968)(3,11.891)(4,7.011)(5,4.585)};

\addplot+[color=red,fill=red,postaction={pattern=vertical lines,pattern color=red!40!white}] coordinates {(1,71.46)(2,3.158)(3,12.224)(4,8.904)(5,4.804)};

\legend{\phantom{0}50 docs/step,100 docs/step,150 docs/step}

\end{axis}
\end{tikzpicture}
\hskip 1pt
\begin{tikzpicture}
\begin{axis}[
    ybar,
    height=3.5cm,
    bar width=7pt,
    title={\hotpotqa{}},
    xlabel={Retrieval Steps/Question},
    xtick={1,2,3,4,5},
    xmin=.5,xmax=5.5,
    legend pos=north east,
    ymajorgrids=true,
    grid style=dashed,
]

\addplot+[color=blue,fill=blue,postaction={pattern=crosshatch dots,pattern color=blue!40!white}] coordinates {(1,0.351)(2,84.56)(3,12.586)(4,1.958)(5,0.540)};

\addplot+[color=green,fill=green,postaction={pattern=horizontal lines,pattern color=green!40!white}] coordinates {(1,0.27)(2,85.82)(3,12.07)(4,1.404)(5,0.432)};

\addplot+[color=red,fill=red,postaction={pattern=vertical lines,pattern color=red!40!white}] coordinates {(1,0.64)(2,86.75)(3,8.90)(4,2.876)(5,0.8237)};

\legend{\phantom{0}50 docs/step,100 docs/step,150 docs/step}

\end{axis}
\end{tikzpicture}
}
\\
\resizebox{.48\textwidth}{!}{
\begin{tikzpicture}
\begin{axis}[
    xlabel={Total Paragraphs Retrieved/Question},
    ylabel={Answer \fone{}},
    height=5cm,
    xmin=-50, xmax=550,
    ymin=57.4, ymax=62.6,
    xtick={0,100,200,300,400,500},
    ytick={58,59,60,61,62},
    legend pos=south east,
    ymajorgrids=true,
    grid style=dashed,
]

\addplot[color=blue,mark=square,]
coordinates  {(50,58.49)(80,58.80)(109,58.64)(132,58.18)(151,58.00)}
node[pos=0,left]{(1)} node[pos=1,right]{(5)};
\addplot[color=green,mark=star,]
coordinates {(100,60.81)(155,61.09)(208,60.72)(250,60.34)(286,60.20)} 
node[pos=0,left]{(1)} node[pos=1,right]{(5)};
\addplot[color=red,mark=triangle,]
coordinates {(150,61.74)(229,61.94)(305,61.63)(364,61.05)(412,60.91)}
node[pos=0,left]{(1)} node[pos=1,right]{(5)};

\legend{\phantom{0}50 docs/step,100 docs/step,150 docs/step}

\end{axis}
\end{tikzpicture}
\hskip 1pt
\begin{tikzpicture}
\begin{axis}[
    xlabel={Total Paragraphs Retrieved/Question},
    height=5cm,
    xmin=50, xmax=400,
    ymin=73.9, ymax=79.1,
    xtick={0,100,200,300,400},
    ytick={74,75,76,77,78,79},
    legend pos=south east,
    ymajorgrids=true,
    grid style=dashed,
]

\addplot[color=blue,mark=square,]
coordinates  {(100,75.42)(107,76.94)(109,77.10)(110,77.11)}
node[pos=0,left]{(2)} node[pos=1,right]{(5)};
    
\addplot[color=green,mark=star,]
coordinates {(200,76.17)(214,77.62)(216,77.76)(218,77.72)}
node[pos=0,left]{(2)} node[pos=1,right]{(5)};

\addplot[color=red,mark=triangle,]
coordinates {(300,76.40)(320,78.18)(328,78.40)(333,78.41)}
node[pos=0,left]{(2)} node[pos=1,right]{(5)};

\legend{\phantom{0}50 docs/step,100 docs/step,150 docs/step}

\end{axis}
\end{tikzpicture}
}
\caption{The retrieval behavior of \irrr{} and its relation to the performance of end-to-end question answering.
Top: The distribution of reasoning path lengths as determined by \irrr{}\@. Bottom: Total number of paragraphs retrieved by \irrr{} vs.\ the end-to-end question answering performance as measured by answer \fone{}.
} \label{fig:retrieval}
\end{figure}
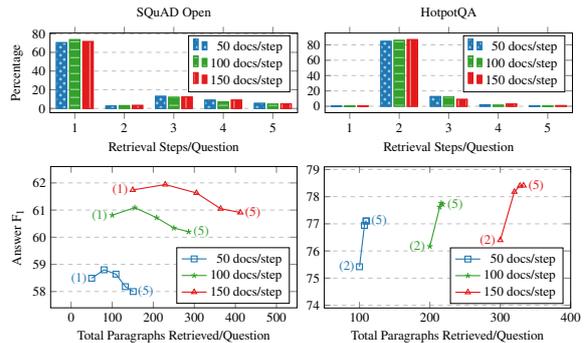

To better understand the behavior of \irrr{} on these benchmarks, we analyze the number of paragraphs retrieved by the model when varying the number of paragraphs retrieved at each reasoning step among $\{50, 100, 150\}$.
As can be seen in Figure \ref{fig:retrieval}, \irrr{} stops its iterative process as soon as all necessary paragraphs to answer the question have been retrieved, effectively reducing the total number of paragraphs retrieved and read by the model compared to always retrieving a fixed number of paragraphs for each question.
Further, we note that the optimal cap for the number of reasoning steps is larger than the number of gold paragraphs necessary to answer the question on each benchmark, which we find is due to \irrr{}'s ability to recover from retrieving and selecting non-gold paragraphs (see the example in Figure~\ref{fig:example_query_prediction}).
Finally, we note that increasing the number of paragraphs retrieved at each reasoning step remains an effective, if computationally expensive, strategy, to improve the end-to-end performance of \irrr{}.
However, the tradeoff between retrieval budget and model performance is more effective than that of previous work (\eg, GRR), and we note that the queries generated by \irrr{} are explainable to humans and can help humans easily control its behavior.

\subsection{Performance on the Unified Benchmark}

\begin{table}
\small
\centering
    \resizebox{.45\textwidth}{!}{
    \begin{tabular}{lcccccc}
        \toprule
        \multirow{2}{2cm}{} & \multicolumn{2}{c}{Dev} && \multicolumn{2}{c}{Test} \\
         \cline{2-3} \cline{5-6}  
         & EM & \fone{} && EM & \fone{} \\ 
         \midrule
         \squad{} Open  & 50.65 & 60.99 && 60.59 & 67.51  \\
         \hotpotqa{}    & 59.01 & 70.33 && 58.61 & 69.86  \\
         3+ hop & --- & --- && 33.02 & 39.59  \\
         \midrule
         Micro-averaged & 54.20 & 64.95 && 58.82 & 67.46 \\
         Macro-averaged & 54.83 & 65.66 && 50.74 & 58.99 \\
         \bottomrule
    \end{tabular}
    }
    \caption{End-to-end question answering performance of \irrr{} on the unified benchmark, evaluated on the 2020 copy of Wikipedia.
    These results are not directly comparable with those in Tables \ref{tab:benchmark_results_squad} and \ref{tab:benchmark_results_hotpot_3hop} because the set of questions and Wikipedia documents differ.}
    \label{tab:main_results}
\end{table}

\begin{table}
    \small
    \centering
    \resizebox{.48\textwidth}{!}{%
    \begin{tabular}{lccc}
    \toprule
    System & \squad{} & \hotpotqa{}\\
    \midrule
    Ours (joint dataset) & %
    58.69 & %
    68.74  \\
    \quad vs. fixed retrieval steps ($K=3$) & %
    31.70 & %
    66.60 \\
    \quad vs. remove \hotpotqa{} / \squad{} data & 54.35 & %
    66.91 \\
    \quad replace ELECTRA w/ BERT\textsubscript{LARGE-WWM} & 57.19 & %
    63.86 \\
    \bottomrule
    \end{tabular}
    }
    \caption{Ablation study of different design choices in \irrr{}, as evaluated by Answer \fone{} on the dev set of the unified benchmark.
    Results differ from those in Table \ref{tab:main_results} because fewer reasoning steps are used (3 vs.\ 5) and fewer paragraphs retrieved at each step (50 vs.\ 150).}
    \label{tab:ablations}
\end{table}

To demonstrate the performance of \irrr{} in a more realistic setting of open-domain QA, we evaluate it on the new, unified benchmark.
As is shown in Table \ref{tab:main_results}, \irrr{}'s performance remains competitive on all questions from different origins in the unified benchmark, despite the difference in reasoning complexity when answering these questions.
The model also generalizes to the 3-hop questions despite having never been trained on them.
We note that the large performance gap between the development and test settings for \squadopen{} questions is due to the fact that test set questions (the original \squad{} dev set) are annotated with multiple human answers, while the dev set ones (originally from the \squad{} training set) are not.

\begin{figure}
    \centering
    \resizebox{0.48\textwidth}{!}{
    \small
    \begin{tabular}{p{4.5em}p{25em}}
    \toprule
    Question & The \textbf{\textcolor{blue}{\underline{Ingerophrynus gollum}}} is \textbf{\textcolor{blue}{\underline{named}}} after a \textbf{\textcolor{blue}{\underline{character}}} in a \textbf{\textcolor{blue}{\underline{book}}} that \textbf{\textcolor{blue}{\underline{sold}}} how many \textbf{\textcolor{blue}{\underline{copies}}}? \\
    \midrule
     Step 1\newline \emph{(Non-Gold)} & Ingerophrynus is a genus of true toads with 12 species. ... In 2007 a new species, ``Ingerophrynus gollum'', was added to this genus. This species is named after the character Gollum created by \textbf{\textcolor{green}{\underline{J. R. R. Tolkien}}}." \\
    Query & \textcolor{blue}{Ingerophrynus gollum book sold copies} \textbf{\textcolor{green}{\underline{J. R. R. Tolkien}}} \\
    \midrule
     Step 2 \newline \emph{(Gold)} & Ingerophrynus gollum (Gollum's toad) is a species of \textbf{\textcolor{purple}{\underline{true}}} toad. ... It is called ``gollum'' with reference of the eponymous character of The \textbf{\textcolor{purple}{\underline{Lord of the Rings}}} by J. R. R. Tolkien. \\
    Query & \textcolor{blue}{Ingerophrynus gollum character book sold copies} \textcolor{green}{J. R. R. Tolkien} \textbf{\textcolor{purple}{\underline{true}} \textcolor{purple}{\underline{Lord of the Rings}}} \\
    \midrule
     Step 3 \newline \emph{(Gold)} & The Lord of the Rings is an epic high fantasy novel written by English author and scholar J. R. R. Tolkien. ...  is one of the best-selling novels ever written, with \textbf{\textcolor{red}{150 million copies}} sold.   \\
     Answer/GT & \textbf{\textcolor{red}{150 million copies}} \\
    \bottomrule
    \end{tabular}
    }
    \caption{An example of \irrr{} answering a question from \hotpotqa{} by generating natural language queries to retrieve paragraphs, then rerank them to compose reasoning paths and read them to predict the answer.
    Here, \irrr{} recovers from an initial retrieval/reranking mistake by retrieving more paragraphs, before arriving at the gold supporting facts and the correct answer.}
    \label{fig:example_query_prediction}
\end{figure}

To better understand the contribution of the various components and techniques we proposed for \irrr{}, we performed ablation studies on the model iterating up to 3 reasoning steps with 50 paragraphs for each step, and present the results in Table \ref{tab:ablations}. %
First of all, we find it is important to allow \irrr{} to dynamically stop retrieving paragraphs to answer the question.
Compared to its fixed-step retrieval counterpart, dynamically stopping \irrr{} improves \fone{} on both \squad{} and \hotpotqa{} questions by 27.0 and 2.1 points respectively (we include further analyses for dynamic stopping in Appendix~\ref{failure_cases}). 
We also find combining SQuAD and HotpotQA datasets beneficial for both datasets in an open-domain setting, and that ELECTRA is an effective alternative to BERT for this task.

\section{Related Work}

The availability of large-scale question answering (QA) datasets has greatly contributed to the research progress on open-domain QA.
\squad{} \citep{rajpurkar2016squad, rajpurkar2018know} is among the first question answering datasets adopted for this purpose by \citet{chen2017reading} to build QA systems over Wikipedia articles.
Similarly, \triviaqa{} \citep{joshi-etal-2017-triviaqa} and Natural Questions \citep{kwiatkowski-etal-2019-natural} feature Wikipedia-based questions that are written by trivia enthusiasts and extracted from Google search queries, respectively.
More recently, \citet{petroni-etal-2021-kilt} presented, KILT, a new benchmark based on Wikipedia where many knowledge-based tasks are evaluated in a unified version of Wikipedia, including open-domain question answering, entity linking, dialogue, etc.
Unlike \beerqa{}, however, single-hop and multi-hop QA are held completely separate during evaluation in KILT, which makes the evaluation of open-domain QA less realistic.
Aside from Wikipedia, researchers have also used news articles \citep{trischler2016newsqa} and search results from the web \citep{dunn2017searchqa, talmor2018web} as the corpus for open-domain QA.

Inspired by the TREC QA challenge,\footnote{\url{https://trec.nist.gov/data/qamain.html}} \citet{chen2017reading} were the first to combine information retrieval systems with accurate neural network-based reading comprehension models for open-domain QA.
Recent work has improved open-domain QA performance by enhancing various components in this retrieve-and-read approach.
While much research focused on improving the reading comprehension model \citep{seo2017bidirectional, clark-gardner-2018-simple}, especially with pretrained langauge models like BERT \citep{devlin2019bert}, researchers have also demonstrated that neural network-based information retrieval systems achieve competitive, if not better, performance compared to traditional IR engines \citep{lee2019latent, khattab2020relevance, guu2020realm, xiong2021answering}.
Aside from the reading comprehension and retrieval components, researchers have also found value from reranking search results \citep{wang2018reinforced} or answer candidates \citep{wang2018evidence, hu2019retrieve}.

While most work focuses on questions that require only a local context of supporting facts to answer, \citet{yang2018hotpotqa} presented \hotpotqa{}, which tests whether open-domain QA systems can generalize to more complex questions that require evidence from multiple documents to answer.
Researchers have explored various techniques to extend retrieve-and-read systems to this problem, including making use of hyperlinks between Wikipedia articles \citep{nie2019revealing, feldman2019multi, zhao2019transformer, asai2020learning, dhingra2020differentiable, zhao2019transformer} and iterative retrieval \citep{talmor2018web, das2019multi, qi2019answering}.
While most previous work on iterative retrieval makes use of neural retrieval systems that directly accept real vectors as input, our work is similar to that of \citet{qi2019answering} in using natural language search queries.
A crucial distinction between our work and previous work on multi-hop open-domain QA, however, is that we don't train models to exclusively answer single-hop or multi-hop questions, but demonstrate that one single set of parameters performs well on both tasks.

\section{Conclusion}

In this paper, we presented Iterative Retriever, Reader, and Reranker (\irrr{}), a system that uses a single model to perform subtasks to answer open-domain questions of arbitrary reasoning steps.
\irrr{} achieves competitive results on standard open-domain QA benchmarks, and establishes a strong baseline on \beerqa{}, the new unified benchmark we present, which features questions with mixed levels of complexity.

\section*{Acknowledgments}

The authors would like to thank the anonymous reviewers for discussions and comments on earlier versions of this paper. 
This research is funded in part by Samsung Electronics Co., Ltd.~and in part by the SAIL-JD Research Initiative.

\bibliographystyle{acl_natbib}
\bibliography{mybib}

\clearpage
\appendix
\section{Data processing} \label{sec:data_processing}

In this section, we describe how we process the English Wikipedia and the \squad{} dataset for training and evaluating \irrr{}.

For the standard benchmarks (\squadopen{} and \hotpotqa{} fullwiki), we use the Wikipedia corpora prepared by \citet{chen2017reading} and \citet{yang2018hotpotqa}, respectively, so that our results are comparable with previous work on these benchmarks.
Specifically, for \squadopen, we use the processed English Wikipedia released by \citet{chen2017reading} which was accessed in 2016, and contains 5,075,182 documents.\footnote{\url{https://github.com/facebookresearch/DrQA}}
For \hotpotqa{}, \citet{yang2018hotpotqa} released a processed set of Wikipedia introductory paragraphs from the English Wikipedia originally accessed in October 2017.\footnote{\url{https://hotpotqa.github.io/wiki-readme.html}}

While it is established that the \squad{} dev set is repurposed as the test set for \squadopen{} for ease of evaluation, most previous work make use of the entire training set during training, and as a result a proper development set for \squadopen{} does not exist.%
\footnote{Thus, if any hyperparameter tuning has been performed, it is usually done to directly maximize the performance on this held-out test set, inflating the performance on this set as a result.}
We therefore resplit the \squad{} training set into a proper development set that is not used during training, and a reduced training set that we use for all of our experiments.
As a result, although \irrr{} is evaluated on the same test set as previous systems, it is likely disadvantaged due to the reduced amount of training data and hyperparameter tuning on this new dev set.
We split the training set by first grouping questions and paragraphs by the Wikipedia entity/title they belong to, then randomly selecting entities to add to the dev set until the dev set contains roughly as many questions as the test set (original \squad{} dev set).
The statistics of our resplit of \squad{} can be found in Table \ref{tab:squad_resplit}.
We make our resplit publicly available to the community at \url{https://beerqa.github.io/}.

For the unified benchmark, we started by processing the English Wikipedia\footnote{Accessed on August 1st, 2020, which contains 6,133,150 articles in total.} with the WikiExtractor \citep{attardi2015wikiextractor}.
We then tokenized this dump and the supporting context used in \squad{} and \hotpotqa{} with Stanford CoreNLP 4.0.0 \citep{manning-EtAl:2014:P14-5} to look for paragraphs in the 2020 Wikipedia dump that might correspond to the context paragraphs in these datasets.
Since many Wikipedia articles have been renamed or removed since, we begin by following Wikipedia redirect links to locate the current title of the corresponding Wikipedia page (\eg, the page \emph{``Madonna (entertainer)''} has been renamed \emph{``Madonna''}).
After the correct Wikipedia article is located, we look for combinations of one to two consecutive paragraphs in the 2020 Wikipedia dump that have high overlap with context paragraphs in these datasets.
We calculate the recall of words and phrases in the original context paragraph (because Wikipedia paragraphs are often expanded with more details), and pick the best combination of paragraphs from the article.
If the best candidate has either more than 66\% unigrams in the original context, or if there is a common subsequence between the two that covers more than 50\% of the original context, we consider the matching successful, and map the answers to the new context paragraphs.
The main causes of mismatches are a) Wikipedia pages that have been permanently removed (due to copyright issues, unable to meet notability standards, etc.); b) significantly edited to improve presentation (see Figure \ref{fig:wiki_changes:madonna}); c) significantly edited because the world has changed (see Figure \ref{fig:wiki_changes:peter_madsen}).

\begin{figure*}
    \small
    \centering
    \subfigure[The Wikipedia page about Madonna, on December 20, 2016 (on the left, which is in the version \squadopen{} used) versus July 31, 2020 (on the right, which is in the version \beerqa{} used).]{
    \parbox{0.47\textwidth}{%
    \begin{framed}
    Madonna Louise Ciccone (born August 16, 1958) is an American singer, songwriter, actress, and \deleted{businesswoman}. \modified{She achieved popularity by pushing the boundaries of lyrical content in mainstream popular music and imagery in her music videos}, \deleted{which became a fixture on MTV.} \modified{Madonna is known for reinventing both her music and image, and for maintaining her autonomy within the recording industry.} Music critics have acclaimed her musical productions, which have generated some controversy. Referred to as the ``Queen of Pop'', Madonna is often cited as an influence by other artists.
    \end{framed}}\ 
    \parbox{0.53\textwidth}{%
    \begin{framed}
    Madonna Louise Ciccone (born August 16, 1958) is an American singer-songwriter, author, actress and \added{record executive}. She has been referred to as the ``Queen of Pop'' \added{since the 1980s}. \modified{Madonna is noted for her continual reinvention and versatility in music production, songwriting, and visual presentation. She has pushed the boundaries of artistic expression in popular culture, while remaining completely in charge of every aspect of her career.} Her works, \added{which incorporate social, political, sexual, and religious themes,} have made a cultural impact which has generated both critical acclaim and controversy. Madonna is often cited as an influence by other artists.
    \end{framed}}
    \label{fig:wiki_changes:madonna}
    }\\
    \subfigure[The Wikipedia page about Peter Madsen, on September 27, 2017 (on the left, which is in the version \hotpotqa{} used) versus July 26, 2020 (on the right, which is in the version \beerqa{} used).]{
    \parbox{0.55\textwidth}{%
    \begin{framed}
    Peter Langkj\ae{}r Madsen (born 12 January 1971) is a Danish \deleted{aerospace engineering enthusiast, ``art engineer'', submarine builder, entrepreneur, co-founder of the non-profit organization Copenhagen Suborbitals, and founder and CEO of RML Spacelab ApS. He was arrested in August 2017 for involvement} in the death of Swedish journalist Kim Wall; \deleted{the investigation is ongoing.}
    \end{framed}}
    \parbox{0.45\textwidth}{%
    \begin{framed}
    Peter Langkj\ae{}r Madsen (]; born 12 January 1971) is a Danish \added{convicted murderer. In April 2018 he was convicted of the 2017} murder of Swedish journalist Kim Wall \added{on board his submarine, UC3 Nautilus, and sentenced to life imprisonment. He had previously been an engineer and entrepreneur.}
    \end{framed}}
    \label{fig:wiki_changes:peter_madsen}
    }
    \caption{Changes in Wikipedia that present challenges in matching them across years. We highlight portions of the text that have been deleted in \deleted{red underlined text}, that have been added in \added{green boldface text}, and that have been significantly paraphrased in \modified{orange italics}, and leave near-verbatim text in the normal font and color.}
    \label{fig:wiki_changes}
\end{figure*}

\begin{table}
    \centering
    \begin{tabular}{cccc}
    \toprule
         Split & Origin & \# Entities & \#QAs \\
    \midrule
        train & train & 387 & 77,087 \\
        dev & train & \phantom{0}55 & 10,512 \\
        test & dev & \phantom{0}48 & 10,570 \\
    \bottomrule
    \end{tabular}
    \caption{Statistics of the resplit \squad{} dataset for proper training and evaluation on the \squadopen{} setting.}
    \label{tab:squad_resplit}
\end{table}

As a result, 20,182/2,146 \squad{} train/dev examples (that is, 17,802/2,380/2,146 train/dev/test examples after data resplit) and 15,806/1,416/1,427 \hotpotqa{} train/dev/fullwiki test examples have been excluded from the unified benchmark.
To understand the data quality after converting \squadopen{} and \hotpotqa{} to the newer version of Wikipedia, we sampled 100 examples from the training split of each dataset.
We find that 6\% of \squad{} questions and 10\% of \hotpotqa{} questions are no longer answerable from their context paragraphs due to edits in Wikipedia or changes in the world, despite the presence of the answer span.
We also find that 43\% of \hotpotqa{} examples contain more than the minimal set of necessary paragraphs to answer the question as a result of the mapping process.

\section{Elasticsearch Setup} \label{sec:search_engine}

We set up Elasticsearch in standard benchmark settings (\squadopen{} and \hotpotqa{} fullwiki) following practices in previous work \citep{chen2017reading, qi2019answering}, with minor modifications to unify these approaches.

Specifically, to reduce the context size for the Transformer encoder in \irrr{} to avoid unnecessary computational cost, we primarily index the individual paragraphs in the English Wikipedia.
To incorporate the broader context from the entire article, as was done by \citet{chen2017reading}, we also index the full text for each Wikipedia article to help with scoring candidate paragraphs.
Each paragraph is associated with the full text of the Wikipedia article it originated from, and the search score is calculated as the summation of two parts: the similarity between query terms and the paragraph text, and the similarity between the query terms and the full text of the article.

For query-paragraph similarity, we use the standard BM25 similarity function \citep{robertson1995okapi} with default hyperparameters ($k_1=1.2, b=0.75$).
For query-article similarity, we find BM25 to be less effective, since the length of these articles overwhelm the similarity score stemming from important rare query terms, which has also been reported in the information retrieval literature \citep{lv2011documents}.
Instead of boosting the term frequenty score as considered by \citet{lv2011documents}, we extend BM25 by taking the square of the IDF term and setting the TF normalization term to zero ($b=0$), which is similar to the TF-IDF implementation by \citet{chen2017reading} that is shown effective for \squadopen{}.

Specifically, given a document $D$ and query $Q$, the score is calculated as
\begin{align}
    \mathrm{score}(D, Q) &= \sum_{i=1}^n \mathrm{IDF}_+^2(q_i) \cdot \frac{f(D, q_i) \cdot (1 + k_1)}{f(D, q_i) + k_1},
\end{align}
where $\mathrm{IDF}_+(q_i) = \max(0, \log((N-n(q_i) + 0.5) / (n(q_i) + 0.5))$, with $N$ denoting the total numberr of documents and $n(q_i)$ the document frequency of query term $q_i$, and $f(q_i, D)$ is the term frequency of query term $q_i$ in document $D$.
We set $k_1 = 1.2$ in all of our experiments.
Intuitively, compared to the standard BM25, this scoring function puts more emphasis on important, rare term overlaps while it is less dampened by document length, making it ideal for an initial sift to find relevant documents for open-domain question answering.

\section{Further Training and Prediction Details} \label{sec:further_training_details}

We include the hyperparameters used to train the \irrr{} model in Table \ref{tab:hyperparams} for reproducibility.

For our experiments using \squad{} for training, we also follow the practice of \citet{asai2020learning} to include the data for \squad{} 2.0 \citep{rajpurkar2018know} as negative examples for the reader component.
Hyperparameters like the prediction threshold of binary classifiers in the query generator are chosen on the development set to optimize end-to-end QA performance.

We also include how we use the reader model's prediction to stop the \irrr{} pipeline for completeness.
Specifically, when the most likely answer is yes or no, the answerability of the reasoning path is the difference between the yes/no logit and the \texttt{NOANSWER} logit.
For reasoning paths that are not answerable, we further train the span classifiers to predict the \clstoken{} token as the ``output span'', and thus we also include the likelihood ratio between the best span and the \clstoken{} span if the positive answer is a span.
Therefore, when the best predicted answer is a span, its answerability score is computed by including the score of the ``\clstoken{} span'' as well, \ie,
\begin{align}
    \mathrm{Answerability}_{\text{span}}(p) &= \mathrm{logit}_{\text{span}} - \mathrm{logit}_{\texttt{NOANSWER}} \nonumber \\
    &+ \frac{\mathrm{logit}_{s}^{\text{start}} - \mathrm{logit}_{\clstoken}^{\text{start}}}{2} \nonumber \\
    &+ \frac{\mathrm{logit}_{e}^{\text{end}} - \mathrm{logit}_{\clstoken}^{\text{end}}}{2},
\end{align}
where $\mathrm{logit}_{\text{span}}$ is the logit of predicting span answers from the 4-way classifier, while $\mathrm{logit}^{\text{start}}$ and $\mathrm{logit}^{\text{end}}$ are logits from the span classifiers for selecting the predicted span from the reasoning path.

\begin{table}
    \centering
    \small
    \begin{tabular}{lc}
        \toprule
        Parameter & Value \\
        \midrule
        Learning rate & $3\times 10^{-5}$ \\
        Batch size & 320 \\
        Iteration & 10,000 \\
        Warming-up & 1,000 \\
        Training tokens & $1.638\times 10^9$\\
        Reranker Candidates & 5 \\
        \bottomrule
    \end{tabular}
    \caption{Hyperparameter setting for \irrr{} training.} 
    \label{tab:hyperparams}
\end{table}

\section{Further Analyses of Model Behavior}

In this section, we perform further analyses and introduce further case studies to demonstrate the behavior of the \irrr{} system.
We start by analyzing the effect of the dynamic stopping criterion for reasoning path retrieval, then move on to the end-to-end performance and leakages in the pipeline, and end with a few examples to demonstrate typical failure modes we have identified that might point to limitations with the data.

\begin{figure}
    \centering
    \resizebox{0.48\textwidth}{!}{
    \small
    \begin{tabular}{p{6em}p{24em}}
    \toprule
    Question & What \textbf{\textcolor{blue}{\underline{team}}} was the \textbf{\textcolor{blue}{\underline{AFC champion}}}? \\
    \midrule
     Step1\newline (Non-Gold) & However, the eventual-AFC Champion \textbf{\textcolor{green}{\underline{Cincinnati Bengals}}}, playing in their first AFC Championship Game, defeated the Chargers 27-7 in what became known as the Freezer Bowl. ... \\
    \midrule
     Step2 \newline (Non-Gold) & Super Bowl XXVII was an American football game between the American Football Conference (AFC) champion \textbf{\textcolor{purple}{\underline{Buffalo Bills}}} and the National Football Conference (NFC) champion Dallas Cowboys to decide the National Football League (NFL) champion for the 1992 season. ... \\
    \midrule
     Gold & Super Bowl 50 was an American football game to determine the champion of the National Football League (NFL) for the 2015 season. The American Football Conference (AFC) champion \textbf{\textcolor{red}{\underline{Denver Broncos}}} defeated the National Football Conference (NFC) champion Carolina Panthers 24-10 to earn their third Super Bowl title. ... \\
    \bottomrule
    \end{tabular}
    }
    \caption{An example where there are false negative answers in Wikipedia for the question from \squadopen{}.}
    \label{fig:squad_failure_case}
\end{figure}

\paragraph{Effect of Dynamic Stopping.}
We begin by studying the effect of using the answerability score as a criterion to stop the iterative retrieval, reading, and reranking process within \irrr{}.
We compare the performance of a model with dynamic stopping to one that is forced to stop at exactly $K$ steps of reasoning, neither more nor fewer, where $K=1, 2, \ldots, 5$.
As can be seen in Table \ref{tab:squad_hotpotqa_adaptive}, \irrr{}'s dynamic stopping criterion based on the answerability score is very effective in achieving good end-to-end question answering performance for questions of arbitrary complexity without having to specify the complexity of questions ahead of time.
On both \squadopen{} and \hotpotqa{}, it achieves competitive, if not superior question answering performance, even without knowing the true number of gold paragraphs necessary to answer each question.

\begin{table}
    \centering
    \small
    \begin{tabular}{lccp{.1em}cc}
    \toprule
    \multirow{2}{3em}{Steps} & \multicolumn{2}{c}{\squadopen{}} && \multicolumn{2}{c}{\hotpotqa{}}\\
    \cline{2-3}\cline{5-6}
    & EM & \fone && EM & \fone
    \\
    \midrule
    Dynamic &  49.92 & 60.91 && \textbf{65.74} & \textbf{78.41} \\
    1 step & \textbf{51.07} & \textbf{61.74} && 13.75 & 18.95 \\
    2 step & 38.74 & 48.61 && 65.12 & 77.75 \\
    3 step & 32.14 & 41.66 && 65.37 & 78.16 \\
    4 step & 29.06 & 38.33 && 63.89 & 76.72 \\
    5 step & 19.53 & 25.86 && 59.86 & 72.79 \\
    \bottomrule
    \end{tabular}
    \caption{\squad{} and \hotpotqa{} performance using adaptive vs.\ fixed-length reasoning paths, as measured by answer exact match (EM) and \fone{}.
    The dynamic stopping criterion employed by \irrr{} achieves comparable performance to its fixed-step counterparts, without knowledge of the true number of gold paragraphs.}
    \label{tab:squad_hotpotqa_adaptive}
\end{table}

Aside from this, we note four interesting findings: (1) the performance of \hotpotqa{} does not peak at two steps of reasoning, but instead is helped by performing a third step of retrieval for the average question; (2) for both datasets, forcing the model to retrieve more paragraphs after a point consistently hurt QA performance; (3) dynamic stopping slightly hurts QA performance on \squadopen{} compared to a fixed number of reasoning steps ($K=1$); (4) when \irrr{} is allowed to select a dynamic stopping criterion for each example independently, the resulting question answering performance is better than a one-size-fits-all solution of applying the same number of reasoning steps to all examples.
While the last confirms the effectiveness of our answerability-based stopping criterion, the cause behind the first three warrants further investigation.
We will present further analyses to shed light on potential causes of these in the remainder of this section.

\paragraph{Case Study for Failure Cases.} \label{failure_cases}

Besides model inaccuracy, one common reason for \irrr{} to fail at finding the correct answer provided with the datasets is the existence of false negatives (see Figure \ref{fig:squad_failure_case} for an example from \squadopen{}).
We estimate that there are about 9\% such cases in the \hotpotqa{} part of the training set, and 26\% in the \squad{} part of the training set.

These false negatives hurt the quality of data generation as well, especially when generating the \squad{} part of the training set. We investigate randomly selected question-context pairs in the training set and find 24\% of our \squad{} training set and 13\% of GRR's \squad{} training set are false negatives. This means our methods find better candidate documents but true answers in those documents become false positives. That results in worse performance for our model when it is trained with only the \squad{} part of training set as shown in Table~\ref{tab:benchmark_results_squad}. 

\section{Three+ Hop Challenge Set Analysis} \label{sec:three_hop}
Although \squadopen{} and \hotpotqa{} probe our model's ability on single and two-hop questions, we lacked insight into the ability of our model to generalize to questions that require three or more reasoning steps/hops, which is more than what our model is trained on. 
Therefore, we built a challenge set comprised of questions that require at least three hops of reasoning to answer (see Table \ref{tab:hops-stats} for a breakdown of the number of documents required to answer each question in the challenge set). While the vast majority of challenge set questions require three documents, questions that require four or more documents are also present, hence the ``Three+ Hop Challenge Set'' name.
Although we intend to use the challenge set for testing only, we will share a few key insights into the question sourcing process, the reasoning types required, and the answer types present.

\begin{table}
\small
\begin{tabular}{p{2.5cm}cccccc}
\toprule
\# of Documents to answer the question & 3 & 4 & 5 & 6 & 7 & 8\\
\midrule
\# of questions & 495 & 17 & 8 & 0 & 9 & 1\\
\bottomrule
\end{tabular}

\caption{Distribution of reasoning steps for questions in Three+ Hop Challenge Set.} \label{tab:hops-stats}
\end{table}

\paragraph{Question Sourcing Process.}
We annotated 530 examples that require three or more paragraphs to be answered on the 2020 Wikipedia dump. 
We developed roughly 50--100 question templates that cover a diverse set of topics, including science, literature, film, music, history, sports, technology, politics, and geography. 
We then annotated approximately ten to twenty examples from each of these question templates to ensure that the resulting challenge set contained a diverse set of topics and questions.

\begin{table}
    \centering
\begin{tabular}{lcc}
\toprule
Reasoning Type    & \%   \\ 
\midrule
Comparison        & 25.6 \\
Bridge-Comparison &  25.3 \\
Bridge            & 49.1 \\
\bottomrule
\end{tabular}

\caption{Reasoning types required for Three+ Hop Challenge Set.}
\label{tab:three_hop_reasoning_types}
\end{table}

\paragraph{Reasoning Types.}
During the annotation process for the challenge set, we recorded the types of reasoning required to answer each question (Table \ref{tab:three_hop_reasoning_types}). 
Roughly half of the questions require chain reasoning (Bridge), where the reader must identify bridge entities that link the question to the first context paragraph, the first context paragraph to the second, and finally the second to the third where the answer can be found. In the case that four or more hops of reasoning are required, this chain of reasoning will extend past the third paragraph to the $n$-th paragraph where the answer can be found.
Additionally, approximately 25\% of the questions require the comparison of three or more entities (Comparison). 
For these questions, the reader needs to retrieve three or more context paragraphs identified in the question that are not directly connected to each other and then compare them on certain aspects specified in the question, similar to the comparison questions in \hotpotqa{}. 
The remaining 25\% of the questions require both chain reasoning and the comparison of two or more entities (Bridge-Comparison). 
For these questions, the reader must first identify a bridge entity that links the question to the first context paragraph. They then must identify two or more entities to compare within the first context paragraph. 
Afterwards, they retrieve context paragraphs for each of the aforementioned entities and compare them on certain aspects specified in the question.

\begin{table}
    \centering
    \resizebox{\linewidth}{!}{%
    \begin{tabular}{lcl}
    \toprule
Answer Type   & \%   & Example(s)                                            \\ 
\midrule
Person        & 29 & Kate Elizabeth Winslet                                \\
Number        & 20 & 388,072, 5.5 million                                  \\
Yes / No      & 15 & ---                                                     \\
Group / Org   & 11 & CNN                                                   \\
Date          & 8 & March 28, 1930                                        \\
Other Proper Noun & 7 & Boeing 747-400                                        \\
Creative Work & 5  & ``California Dreams''                                   \\
Location      & 4  & New York City                                         \\
Common Noun   & 1    & comedy-drama                                         \\
\bottomrule
\end{tabular}
}
\caption{Types of answers in Three+ Hop Challenge Set. 
These statistics are based on 100 randomly sampled examples.}
\label{tab:three_hop_answer_types}
\end{table} 

\paragraph{Answer Types.}
We also analyze the types of answers present in the challenge set.
As shown in Table \ref{tab:three_hop_answer_types}, the challenge set features a diverse set of answers. 
We find that roughly half of the questions ask about people (29\%) and numeric quantities (20\%).  
Additionally, we find a considerable number of questions that require a yes or no answer (15\%), ask about groups or organizations (11\%), dates (8\%), and other proper nouns (7\%). The challenge set also contains a non-negligible amount of questions that ask about creative works (5\%), locations (4\%), and common nouns (1\%).

\end{document}